\documentclass[10pt,twocolumn,letterpaper]{article}

\usepackage{multirow}
\usepackage{wacv}              
\usepackage[accsupp]{axessibility}
\usepackage{graphicx}
\usepackage{amsmath}
\usepackage{amssymb}
\usepackage{booktabs}
\usepackage{comment}
\usepackage{float}
\usepackage{graphicx}
\usepackage{pifont}
\usepackage{caption}

\usepackage[pagebackref,breaklinks,colorlinks]{hyperref}

\usepackage[capitalize]{cleveref}
\crefname{section}{Sec.}{Secs.}
\Crefname{section}{Section}{Sections}
\Crefname{table}{Table}{Tables}
\crefname{table}{Tab.}{Tabs.}
\newcommand{\cmark}{\ding{51}} 
\newcommand{\xmark}{\ding{55}} 

\def\wacvPaperID{768} 
\def\confName{WACV}
\def\confYear{2026}

\begin{document}

\title{VIZOR: Viewpoint-Invariant Zero-Shot Scene Graph Generation for 3D Scene Reasoning}


\author{%
Vivek Madhavaram$^{1*}$\hspace{0.7cm}
Vartika Sengar$^{2*}$\hspace{0.7cm}
Arkadipta De$^{2*\dagger}$\hspace{0.7cm}
Charu Sharma$^{1}$\hspace{0.7cm}\\
\textsuperscript{1}Machine Learning Lab, IIIT Hyderabad, India\hspace{0.7cm} 
\textsuperscript{2}Fujitsu Research India, Bangalore\\
{\tt\small vivekvardhan.m@research.iiit.ac.in,}
{\tt\small vartika.sengar@fujitsu.com,}
{\tt\small charu.sharma@iiit.ac.in}}

\maketitle 
\def\thefootnote{*}\footnotetext{Equal Contribution}
\def\thefootnote{\textdagger}\footnotetext{Current Affiliation: IBM Research India (Arkadipta.De@ibm.com)}
\begin{abstract}

Scene understanding and reasoning has been a fundamental problem in 3D computer vision, requiring models to identify objects, their properties, and spatial or comparative relationships among the objects. Existing approaches enable this by creating scene graphs using multiple inputs such as 2D images, depth maps, object labels, and annotated relationships from specific reference view. However, these methods often struggle with generalization and produce inaccurate spatial relationships like ``left/right", which become inconsistent across different viewpoints. To address these limitations, we propose Viewpoint-Invariant ZerO-shot scene graph generation for 3D scene Reasoning (VIZOR). VIZOR is a training-free, end-to-end framework that constructs dense, viewpoint-invariant 3D scene graphs directly from raw 3D scenes. The generated scene graph is unambiguous, as spatial relationships are defined relative to each object’s front-facing direction, making them consistent regardless of the reference view. Furthermore, it infers open-vocabulary relationships that describe spatial and proximity relationships among scene objects without requiring annotated training data. We conduct extensive quantitative and qualitative evaluations to assess the effectiveness of VIZOR in scene graph generation and downstream tasks, such as query-based object grounding. VIZOR outperforms state-of-the-art methods, showing clear improvements in scene graph generation and achieving 22\% and 4.81\% gains in zero-shot grounding accuracy on the Replica and Nr3D datasets, respectively. Project page: \url{https://vivekmadhavaram.github.io/vizor/}
\end{abstract}  
\vspace{-0.2cm}
\section{Introduction}
\label{sec:intro}
 
Scene understanding and reasoning is a crucial step for capturing the complete information about the structure of an environment, enabling a wide range of downstream tasks. These downstream tasks include scene captioning \cite{sc1, sc2, sc3, sc4, sc5}, robotic navigation \cite{rn1, rn2, rn3, rn4, rn5, rn6, rana2023sayplan}, scene manipulation \cite{sm1, sm2, sm3, sm4}, change detection \cite{looper20233d}, visual question answering \cite{vqa1, vqa2, vqa3, vqa4} to mention a few. An effective approach to scene understanding involves constructing a scene graph, a structured representation where objects are modeled as nodes, object features as attributes, and relationships among them as edges. An accurate scene graph enables machines to infer spatial and semantic relationships among objects, providing a strong foundation for cognitive perception in various autonomous systems.




\begin{figure}[t]
    \centering
    \includegraphics[width=\columnwidth, trim=0cm 5.8cm 0cm 0cm]{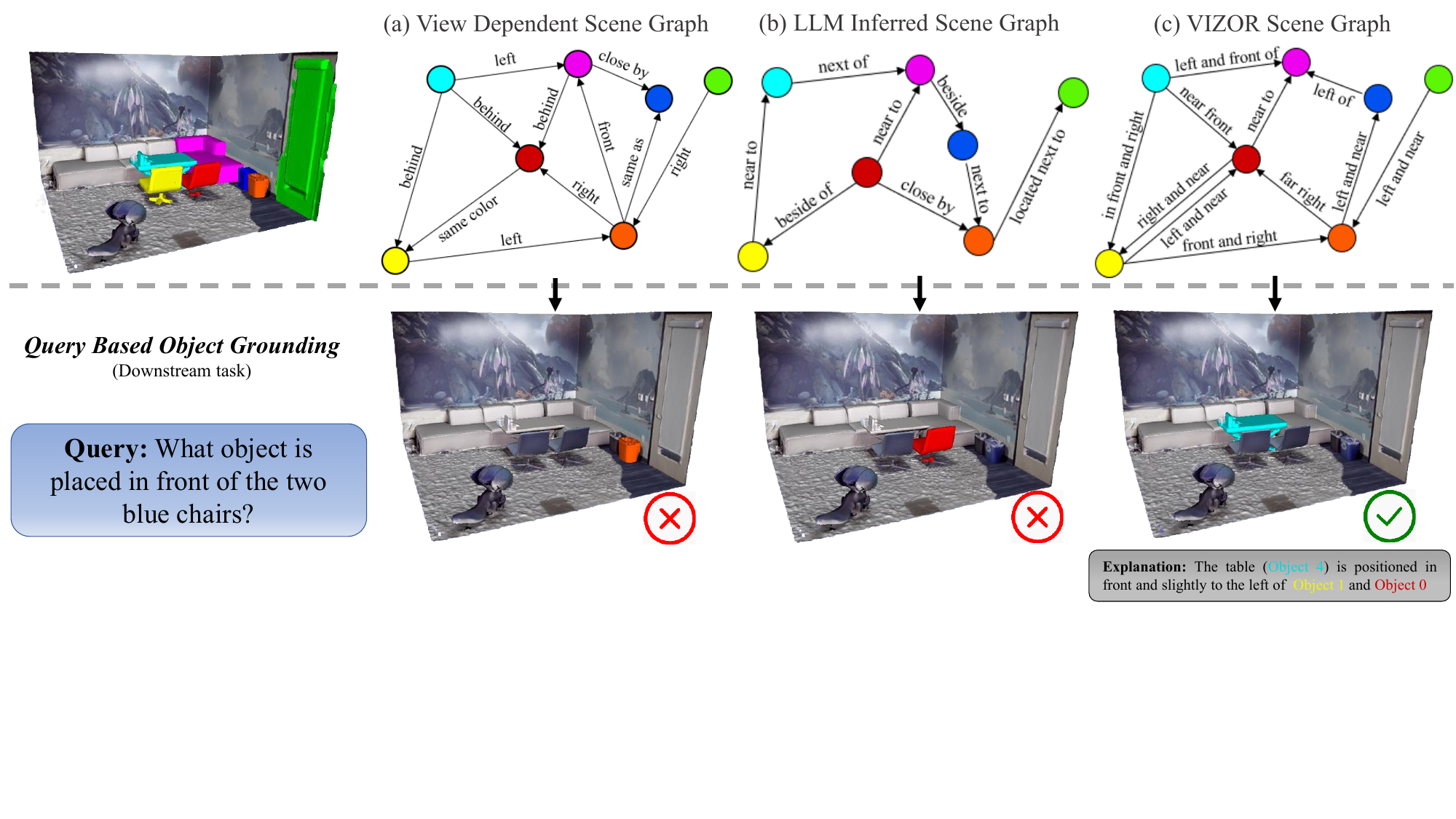}
    \caption{\textbf{Overview:} We present the advantage of view-invariant scene graphs generated by VIZOR, using object grounding as a downstream task. For a given complex query, view-invariant scene graphs (c) capture scene layout more accurately than (a) view-dependent scene graphs, which rely on annotator viewpoint, and (b) LLM-inferred graphs, which often produce overly generic relations.}
    \vspace{-0.5cm}
    \label{fig:teaser}
\end{figure}

Existing approaches \cite{koch2024open3dsg, lv2024sgformer, zhong2021learning, liu2022explore}, for scene graph generation, predominantly rely on supervised learning, where models are trained to predict object relationships using pre-annotated ground truth datasets. However, the availability of large-scale, dense and well-annotated 3D scene graph datasets is highly limited, restricting these approaches to specific scene types and makes model training infeasible in many real-world use cases. Furthermore, datasets such as 3DSSG \cite{3DSSG2020}, annotate spatial and proximity relationships relative to a specific reference view, meaning relationships like ``left/right" or ``front" are defined based on the annotator’s perspective. An example is shown in Figure \ref{fig:teaser}(a). As a result, these scene graphs become inconsistent when the viewpoint changes, limiting their generalization and applicability to different perspectives and practical applications. 

To eliminate the dependency on annotated datasets, recent approaches (e.g., \cite{gu2024conceptgraphs}) have started exploring the use of Large Language Models (LLMs) to infer object relationships based on common-sense knowledge. While LLMs can generate plausible relations, they lack scene-specific contextual grounding, often producing generic relations. For instance, consider Figure \ref{fig:teaser}(b): two chairs are positioned around a table at the same distance, but their spatial positions differ. The relation inferred between the table and these chairs is ``near to". However, using only this relation for grounding one chair w.r.t the table introduces ambiguity, since both chairs satisfy the condition equally, and it's unclear which is being referred to. Incorporating directional information like ``left/right'', defined relative to each object’s front-facing direction, resolves the ambiguity.

Apart from dataset limitations, few existing methods \cite{linok2024barequeriesopenvocabularyobject} rely on a fixed coordinate system considering (x, y, z) values to define spatial relationships, which can lead to inconsistencies. Typically, one axis is assigned to determine each directional relation. For example, x-axis is used for ``front/behind", y-axis for ``left/right", and z-axis for ``above/below". However, such annotations ignore object orientation, leading to incorrect relations that fail to perform reasoning. Let us assume, for scene in Figure \ref{fig:teaser}, if ``front" is inferred solely based on increasing x-coordinate values, the system might incorrectly return ``recycle bin" as the object in front of the chairs, simply because it has a greater x-coordinate than that of chairs. 
However, based on the actual orientation of the chairs, the correct answer should be ``table". 
This discrepancy arises because axis-based annotation does not account for object-facing direction, leading to semantically incorrect inferences.

To address all the discussed limitations, VIZOR efficiently generates 3D scene graphs, as in Figure \ref{fig:teaser}(c), in a training-free manner using only the input 3D scene. At the core of our approach is a novel object-centric relational framework. VIZOR first detects objects in the scene and determines their front directions. This step is crucial because, it allows spatial relationships to be inferred \textit{relative to how each object is oriented}, rather than relying on the \textit{viewpoint of an annotator}. This ensures that spatial and functional relationships remain consistent regardless of observer position, a key limitation in prior works. 
Using the estimated front directions and object centroids, VIZOR computes bidirectional relationships between object pairs, resulting in a dense and semantically coherent scene graph. Additionally, VIZOR aggregates multi-view object renderings to extract rich object properties, such as appearance, geometry, functionality, and color. These properties are incorporated as node attributes in the graph, enhancing its expressiveness and supporting downstream tasks.

\textbf{Our Contributions:} To the best of our knowledge, we are the first to introduce a zero-shot method for view-independent 3D scene graph generation with relations from the object's perspective. Our contributions are as follows: $i)$ We introduce a training-free framework that constructs structured 3D scene graphs directly from scene meshes, eliminating the need for additional information like RGB-D sequences or scene graph annotated datasets. $ii)$ We introduce an object-centric relation generation mechanism that models spatial relations from the intrinsic viewpoint of each object, making these relations invariant to the observer’s viewpoint. $iii)$ We construct detailed 3D scene graphs where objects serve as nodes, enriched with aggregated attributes and edges represent dense open-vocabulary relations between object pairs. This enables open-ended and fine-grained reasoning over the 3D scene. $iv)$ We validate the effectiveness of VIZOR through extensive qualitative and quantitative evaluations and demonstrate its utility in downstream tasks such as text-based open vocabulary object grounding using generated scene graphs.

\section{Related Works}
\label{sec:related}

\noindent\textbf{3D Scene Graph Generation:} Early work 
introduced hierarchical structures combining buildings, rooms, objects, and cameras \cite{armeni20193d}, later extended to large-scale environments \cite{hughes2022hydra, rosinol20203d, rosinol2021kimera}. Other methods infer
local semantic inter-object relationships \cite{koch2024lang3dsg, wald2020learning, zhang2021exploiting, zhang2021knowledge}. 
Recent approaches \cite{koch2024open3dsg, lv2024sgformer, zhong2021learning, liu2022explore} are fully supervised, relying on annotated graphs such as 3DSSG~\cite{3DSSG2020}.
However, these annotations are limited by their dependence on specific observer's perspective and include sparse relations, which restricts generalization. Models trained on such data typically produce scene graphs valid only from that specific viewpoint, motivating the need for view-invariant 3D scene graph generation.\\
\textbf{Open Vocabulary 3D Scene Graph Generation:} Open-Vocabulary
methods aim to infer diverse, unconstrained relations
beyond fixed label sets.
ConceptGraphs \cite{gu2024conceptgraphs} leverages
2D vision-language-models like GPT-4 \cite{achiam2023gpt} and Llava \cite{liu2023visual} to build queryable 3D scene graphs with captions.
However, it
relies on limited frames containing a subset of objects and lacks contextual grounding, often producing generic relations
disconnected from full 3D scene structure.\\
\textbf{LLMs and VLMs for Scene Understanding:} 
LLMs and Vision-Language Models (VLMs) have advanced instruction-driven 3D scene understanding and synthesis. 
InstructScene\cite{lin2023instructscene} 
combines semantic graph prior 
with VLM-based layout decoding for controllable 
synthesis, while Layout-GPT\cite{feng2023layoutgpt} 
composes in-context 
demonstrations for plausible layout generation.
These methods highlight 
LLMs ability to capture complex scene relationships.

\noindent\textbf{2D and 3D Object Grounding:} 
Grounding 
localized objects in 2D or 3D data based on textual queries. One-stage methods jointly embed text and visual features for direct bounding-box regression \cite{kamath2021mdetr, luo20223d}, while two-stage approaches follow \textit{detect-and-match} paradigm \cite{yu2016modeling, chen2020scanrefer, achlioptas2020referit3d, yu2018mattnet}, generating object proposals and then matching them to language. Graph-based methods \cite{velivckovic2018graph, wang2019dynamic} 
infer spatial relations by connecting objects into
relational graphs \cite{achlioptas2020referit3d, huang2021text, yuan2021instancerefer}. 
Transformer-based approaches \cite{he2021transrefer3d, yang2021sat, roh2022languagerefer, huang2022multi, luo20223d} further improve grounding.
Recent zero-shot method ZS3DVG \cite{yuan2024visual}, VLMGrounder \cite{xuvlm} removes training via LLMs/VLMs and visual programming, but lacks global scene structure and spatial reasoning. VPP-Net \cite{vppnet} investigates the significance of viewpoint information 
and proposes uniform object representation loss to encourage viewpoint invariance
in learned object representations. 3DTRL \cite{shang2022learning} proposes a token representation layer which estimates the 3D positional information of the visual tokens and leverages it for learning viewpoint-agnostic representations.

\noindent Differentiating from prior methods, we propose a training-free, viewpoint-invariant approach for generating dense and expressive 3D scene graphs. Our graphs support open-vocabulary object classes and relations and enable complex spatial and semantic reasoning for downstream tasks without relying on annotations or fixed observer viewpoints.

\section{Proposed Method}
\label{sec:proposed}

\begin{figure*}[t]

    \centering
    \includegraphics[width=0.95\linewidth, trim= 0cm 2cm 0cm 1cm]{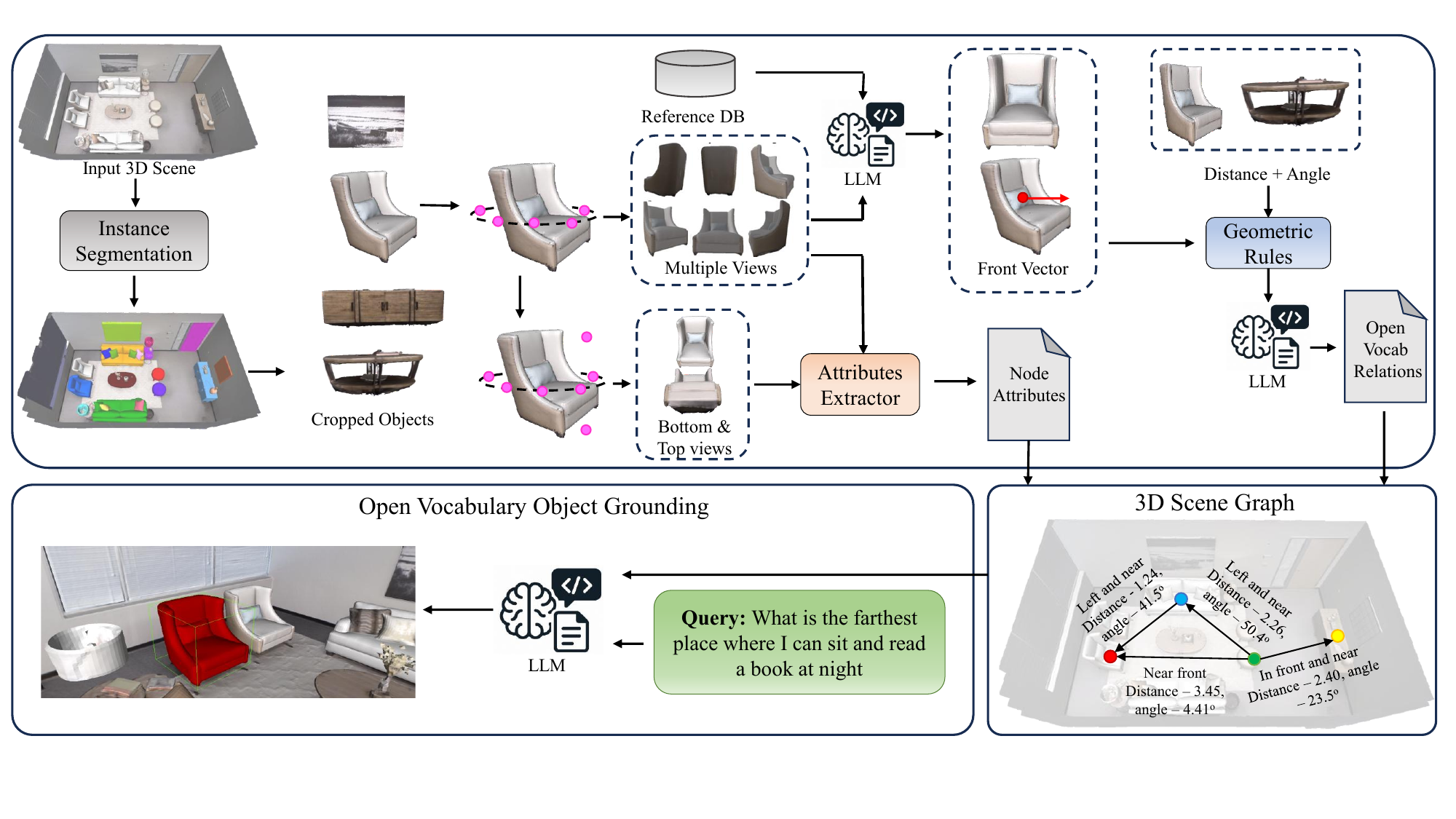}%

    \caption{\textbf{Overall architecture of VIZOR}. It has three major components: Object Segmentation \& Front-direction prediction module (Sec 3.1), object attribute extraction module (Sec 3.2), and Relationship extraction and enrichment module (Sec 3.3). Given only an input 3D scene mesh data, it generates a 3D view-invariant scene graph that can be directly applied to a variety of tasks \textit{viz.} query based open vocabulary object grounding, complex scene understanding etc.}
    \label{fig:VIZOR_main}
    \vspace{-0.5cm}
\end{figure*}

This section presents our proposed method for generating 3D scene graphs. VIZOR takes a 3D scene mesh as input and generates a graph with objects in the scene as nodes and relations between them as edges. Each node in the scene graph has a set of visual and geometric attributes like color, functionality, geometry and caption. An edge between two nodes represents open-vocabulary relations. 

Formally, given an input 3D scene mesh $\mathcal{S} = (\mathcal{P}, \mathcal{F})$ where $\mathcal{P} \in \mathbb{R}^{M \times 6}$ is the set of colored point cloud vertices and $\mathcal{F}$ is the set of faces, 
VIZOR constructs a 3D scene graph, $\mathcal{G} = (\mathcal{V}, \mathcal{E})$ where $\mathcal{V}$ is set of object nodes and $\mathcal{E}$ is the set of relationships (directed edges). The set of nodes is represented as $\mathcal{V} = \{ v_i \mid v_i = (o_i, c_i, A_i) \}, $ where $v_i$ represents an object node with object ID $o_i$, class label $c_i$, and a set of attributes $A_i$, $\quad A_i = \{ a_1, a_2, \dots, a_m \}$. The set of relationships is given by $\mathcal{E} = \{ e_{ij} \mid e_{ij} = (i, j, r_{ij}, d_{ij}, \theta_{ij}) \}$ where $e_{ij}$ represents an edge from $v_i$ to $v_j$ with a relationship label $r_{ij}$, distance between centroids $d_{ij}$, and angle $\theta_{ij}$, an orientation of $v_i$ with respect to front vector of object $v_j$. The overall method to generate the scene graph $\mathcal{G}$ is shown in Figure \ref{fig:VIZOR_main}. It comprises of three main stages: 1) Object segmentation and finding the front direction of each object, 2) Attribute generation for each object, and 3) Predicting relations between pairs of objects.

\subsection{Object Segmentation \& Front Direction Estimation}
\label{sec:frontdir}
To construct a view invariant 3D scene graph, we begin by identifying object instances in the scene. Our framework supports any off-the-shelf 3D segmentation method (e.g., SAM3D~\cite{yang2023sam3d}, Mask3D~\cite{Schult23ICRA}, OpenMask3D~\cite{takmaz2023openmaskd}) to determine objects that serve as nodes. In our implementation, we use Mask3D for instance-level segmentation, producing segmented object meshes $\mathcal{M}_i$, each centered at $\mathbf{c}_{\text{obj}} \in \mathbb{R}^3$. Following segmentation, the next objective is to estimate its \textit{front-facing direction},
which is crucial for constructing a \textit{non observer-perspective, invariant} scene graph. Here, the relations are defined over subject-object pairs, where the subject is the entity from which the relation originates, and the object is the one to which it applies. To accurately compute these spatial relationships, both the front direction and positions of the involved entities are required.

To infer front direction, we build a reference database $\mathcal{R}$ containing one front-view image per class across the 200 Mask3D-recognizable categories~\cite{Schult23ICRA}. For each class, we generate a 3D object using Shap-E~\cite{jun2023shap}, a transformer-based generative model. Since generated object can be fully controlled, it is rotated by a fixed angle such that it faces the camera, and the resulting image is saved as class-specific front-view reference. Front views of missing objects in the reference database can be generated during inference.

Next, for a segmented object $\mathcal{M}_i$, we simulate a circular camera rig of radius $r$ in the plane parallel to XY-plane, centered at object centroid $\mathbf{c}_{\text{obj}} \in \mathbb{R}^3$. A camera is moved along this rig at $N$ uniformly spaced azimuth angles $\theta_k$ to capture multi-view renderings of the object:
\vspace{-0.15cm}
\begin{equation}
\resizebox{0.8\columnwidth}{!}{$
\mathbf{c}_{\text{cam}}^{(k)} = \mathbf{c}_{\text{obj}} + r 
\begin{bmatrix}
\cos\theta_k \\
\sin\theta_k \\
0
\end{bmatrix}, 
\quad \theta_k = \frac{2\pi k}{N}, 
\quad k = 1,\dots,N.
$}
\label{eq:camera_positions}
\end{equation}

The resulting multi-view images $\{I^{(k)}\}$ are compared against the class-specific reference image $I_{\text{ref}}$ (retrieved using class label from $\mathcal{R}$) by a multi-modal LLM (MLLM), which is prompted to identify the front view and confidence scores of each view to $I_{\text{ref}}$ for being the front view. The response generated by MLLM is used to retrieve best matching front view camera position.
The selected camera position $\mathbf{c}_{\text{cam}}^{(k^*)}$ defines the object’s front direction unit vector:
\vspace{-0.2cm}
\begin{equation}
\mathbf{\hat f}_{obj} = \frac{\mathbf{c}_{\text{cam}}^{(k^*)} - \mathbf{c}_{\text{obj}}}
                   {\left\| \mathbf{c}_{\text{cam}}^{(k^*)} - \mathbf{c}_{\text{obj}} \right\|}
\label{eq:front_direction}
\end{equation}

Given object centroid $\mathbf{c}_{\text{obj}}$, we compute relative vector:
\vspace{-0.3cm}
\begin{equation}
\mathbf{r}_{\text{obj} \rightarrow \text{cam}} = \mathbf{c}_{\text{cam}}^{(k^*)} - \mathbf{c}_{\text{obj}}
\label{eq:relative_vector}
\end{equation}

\textbf{Resolving Ambiguous Front View:} For symmetric objects where the front direction is ambiguous (confidence score of more than one view is greater than a front-view threshold), we resolve the ambiguity by selecting the direction that is most aligned with the center of the scene. Let $\{\mathbf{r}_{\text{obj} \rightarrow \text{cam}}^{(i)}\}$ denote the set of candidate front direction vectors for object $\mathcal{M}_i$, derived from multiple object-camera viewpoint configurations. We select the vector forming the smallest angle with the vector pointing from the object centroid $\mathbf{c}_{\text{obj}}$ to the scene center $\mathbf{c}_{\text{scene}}$:

\begin{equation}
\resizebox{0.9\columnwidth}{!}{$
\mathbf{r}_{\text{obj} \rightarrow \text{cam}} = \arg\min_i \; \cos^{-1} \left( 
\frac{ \mathbf{r}_{\text{obj} \rightarrow \text{cam}}^{(i)} \cdot (\mathbf{c}_{\text{scene}} - \mathbf{c}_{\text{obj}}) }
{ \left\| \mathbf{r}_{\text{obj} \rightarrow \text{cam}}^{(i)} \right\| \cdot \left\| \mathbf{c}_{\text{scene}} - \mathbf{c}_{\text{obj}} \right\| } 
\right)
$}
\label{eq:symmetric_front_selection}
\end{equation}

\subsection{Attribute Extraction for Graph Nodes}
\label{sec:attributes}
Each node in our scene graph is represented by a set of descriptive attributes obtained by Attributes Extractor from Figure \ref{fig:VIZOR_main}. These attributes encapsulate various object properties, including physical characteristics (e.g., color, geometry), functionality, style, and semantic descriptions. 

With segmented objects and estimated front directions available from Sec \ref{sec:frontdir}, we generate additional views to support attribute extraction. Specifically, we position a virtual camera in the object's front direction and render a top view by moving camera along positive Z-axis in such a way that its elevation is increased and rotating it towards the object centroid. Similarly, bottom view is captured by moving camera along the negative Z axis. In addition to these, we select every alternate view from the multi-view renderings used in Sec \ref{sec:frontdir}. All of these views are passed to pre-trained MLLM,  along with a structured prompt that specifies the types of attributes to extract.
For each view, the model outputs a structured JSON containing various object properties, including color, geometry, functionality, structural details, and a descriptive caption. 
To obtain a unified representation of the object, we aggregate all per-view JSON outputs by prompting an LLM to consolidate them into a single summary JSON. This final representation captures complementary cues observed across different viewpoints.

Each node in the scene graph is thus defined by the segmented object $\mathcal{M}_i$ and a consolidated attribute set $A_i$. This representation enables a richer understanding of the scene, allowing for downstream reasoning tasks such as relationship inference, object classification and grounding.

\subsection{Computing Edges between Pairs of Objects}
\label{sec:compedges}
The final and most crucial step in constructing a scene graph is establishing edges, which encode the spatial and proximity relationships between subject-object pairs in the scene. Rather than relying on a fixed set of pre-defined relationships, we predict open vocabulary relationships to capture spatial relations in detailed way. Each edge represents a directed connection from a subject node to an object node, indicating spatial configuration of the subject relative to the object in scene layout.

To infer these relations, we leverage: the front direction vector of the object $\hat{f}_{obj}$ (Sec \ref{sec:frontdir}), and the centroids of both the subject $c_{obj_{i}}$ and the object $c_{obj_{j}}$. We compute the euclidean \textit{distance} $d_{ij} = \| c_{obj_{i}} - c_{obj_{j}} \|$ and the \textit{angle} $\theta_{ij}$ between the object's front direction $\vec f_j$ and the vector pointing from the object to the subject, defined as $\vec{v}_{ji} = c_{{obj}_i} - c_{obj_{j}}$. This angle captures the relative orientation of the subject with respect to the object’s viewing direction.

We define a set of geometric rules based on these features to predict the initial set of spatial relationships \textit{viz.} ``in front of", ``behind", ``left", ``right" etc. More complex spatial relations like ``right and above", ``left and behind", ``in front of and to the right" etc., are defined through conjunctions of geometric rules based on the angle between the centroids of the object and subject. Vertical and contact relationships like ``above", ``below", ``on", etc., are determined by analyzing bounding box extents and distance between centroids. 

For each segmented object, we iteratively assign it the role of the object, and compute its relationships with all other subjects. This process continues until every segment in the scene has been treated as the object once. The result is a dense set of directed edges capturing all pairwise spatial interactions. To enhance semantic richness, the full set of predicted relationships, along with the object-level features (i.e., position, bounding box extent, and other extracted properties as described in Sec \ref{sec:attributes}), are passed to LLM. The LLM generates open-vocabulary descriptions that combine spatial and semantic cues, including proximity terms like ``near'' and ``far''. Detailed generation of these relations and prompts are mentioned in supplementary.

\begin{figure}[h!]
\centering
    \vspace{-0.2cm}
    \includegraphics[width=0.7\columnwidth, trim = 2cm 20cm 2cm 0.5cm]{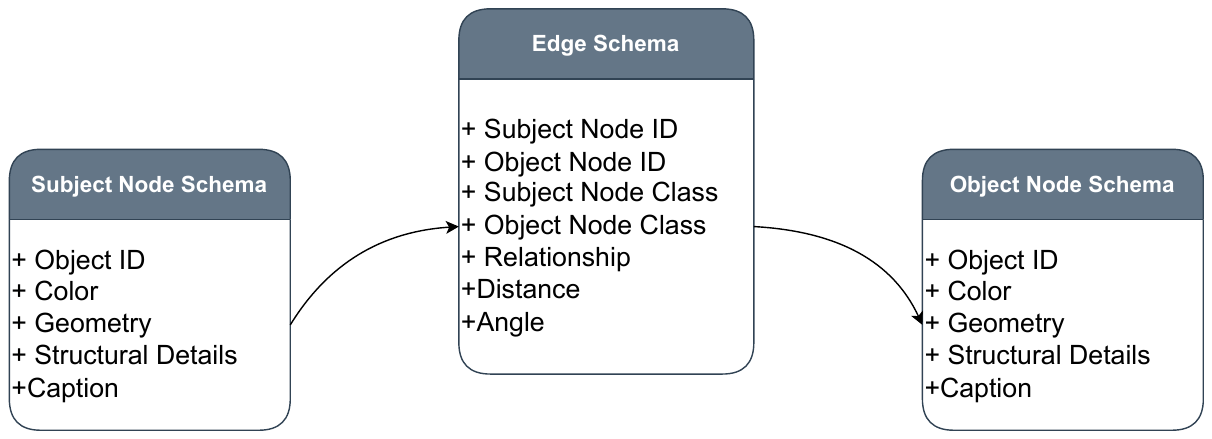}
    \caption{\textbf{Scene Graph Schema}}
    \label{fig:scene-graph-schema}
    \vspace{-0.2cm}
\end{figure}

Finally, each edge in the scene graph is a tuple consisting of the subject and object information, relationship $r_{ij}$, distance between the pairs $d_{ij}$, and angle between the pairs $\theta_{ij}$. The schema of nodes and edges are shown in Figure \ref{fig:scene-graph-schema}.
\section{Experiments}
\label{sec:exp}
 
\subsection{Datasets}
\label{sec:dataset}
\textbf{Replica:} 
Replica \cite{replica19arxiv} is a synthetic 3D dataset containing 18 highly photo-realistic 3D indoor scene reconstructions of rooms, offices, hotels, and apartments. Each scene includes dense mesh, high-resolution, high-dynamic-range (HDR) textures, per-primitive semantic and instance annotations, and planar mirror and glass reflectors. We use the rooms and office scenes for our experiments.

\textbf{Nr3D:} The Nr3D dataset \cite{achlioptas2020referit3d} provides natural language referential annotations for the ScanNet 3D scene dataset \cite{dai2017scannet}, containing 45,503 utterances across 707 indoor scenes. Each scene includes up to six distractors (objects that belong to the same class as the target) and multiple utterances that ground an object in the scene. The utterances are categorized by difficulty (easy and hard) and by dependence on viewpoint (view-dependent and view-independent).

\textbf{3DSSG:} 3DSSG \cite{3DSSG2020}, based on 3RScan \cite{Wald_2020_3RScan}, contains 3D scenes annotated with over 1.6 million spatial and functional relationships represented as scene graphs, where objects are nodes and relations (spatial/semantic) are edges. In this work, we use a few randomly sampled scenes from 3DSSG for qualitative analysis to visualize and interpret our model's predictions.


\subsection{Scene Graph Construction}
We evaluate scene graph construction on the Replica dataset~\cite{replica19arxiv}, following the protocol in \cite{gu2024conceptgraphs}, by selecting 8 scenes (room0 to room2 and office0 to office4). Given the open-vocabulary nature of generated scene graphs and the subjectivity involved in its evaluation, we assess by involving human evaluators. 3 annotators evaluate each node and edge. We compute \textit{node precision} as the fraction of nodes for which at least 2 of 3 human evaluators deem the node caption to be correct. Similarly, an edge (relation between object pairs) is deemed correct if 2 out of 3 evaluators agree on its validity, forming our \textit{edge precision}. The total number of nodes for evaluation is $134$, and we also report the \textit{total number of predicted relationships} in the last column. To evaluate our \textit{front-view prediction} (Sec \ref{sec:frontdir}), we compute the fraction of predicted-front views of segmented objects for which at least 2 out of 3 evaluators deem it to be front-view. The results are shown in Table \ref{tab:sg_result}.
\begin{table}[h!]
\centering
\resizebox{\columnwidth}{!}{%
\begin{tabular}{c|cccc|c|c|c} \hline 
\multirow{3}{*}{Scene} & \multicolumn{4}{c|}{Node Prec.} & Front-View Prec. & Edge Prec. & \# Edges \\ \cline{2-8}   
& CG & CG-D & \begin{tabular}[c]{@{}c@{}}VIZOR\\(LLaMA)\end{tabular} 
& \begin{tabular}[c]{@{}c@{}}VIZOR\\(GPT)\end{tabular} 
& \begin{tabular}[c]{@{}c@{}}VIZOR\end{tabular} 
& \begin{tabular}[c]{@{}c}VIZOR\end{tabular} 
& VIZOR \\ \hline 
room0    & 0.78 & 0.56 & 0.71 & \textbf{0.91} & 0.67 & 0.66 & 420 \\
room1    & 0.77 & 0.70 & 0.61 & \textbf{0.94} & 0.72 & 0.71 & 306 \\
room2    & 0.66 & 0.54 & 0.60 & \textbf{1.00} & 0.80 & 0.73 & 210 \\
office0  & \textbf{0.65} & 0.59 & 0.55 & 0.63 & 0.82 & 0.87 & 110 \\
office1  & 0.65 & 0.49 & 0.58 & \textbf{0.83} & 0.75 & 0.71 & 132 \\
office2  & 0.75 & 0.67 & 0.65 & \textbf{0.90} & 0.55 & 0.64 & 380 \\
office3  & 0.68 & 0.71 & 0.77 & \textbf{0.88} & 0.58 & 0.55 & 272 \\
office4  & -    & -    & 0.70 & \textbf{0.85} & 0.47 & 0.52 & 380 \\ \hline
Average  & 0.71 & 0.61 & 0.65 & \textbf{0.88} & 0.65 & 0.67 & 276 \\ \hline
\end{tabular}%
}
\caption{\textbf{Accuracy of constructed scene graphs on Replica:} Node Prec.: Accuracy of each node caption; Front-View Prec.: Accuracy of predicting the front-view of segmented object correctly, Edge Prec.: Accuracy of each estimated spatial relationship, \# Edges: Number of edges or relations predicted. VIZOR indicates results for both LLaMA and GPT variants. All values are evaluated by human evaluators.}
\label{tab:sg_result}
\end{table}
\vspace{-0.4cm}

We compare our method with two variants of  ConceptGraphs\cite{gu2024conceptgraphs}: the original implementation (CG) and ConceptGraphs-Detector (CG-D) which integrates a RAM \cite{zhang2024recognize} for object identification in the image and Grounding DINO \cite{liu2024grounding} for open-vocabulary object detection. As shown in Table~\ref{tab:sg_result}, VIZOR (GPT) achieves the highest node precision with a (\textbf{17\%} improvement) over the state-of-the-art \cite{gu2024conceptgraphs}. Our method also predicts significantly more edges, demonstrating a richer graph representation.

\subsection{Object Grounding based on Text Queries} \label{sec:obj_retrive}
We evaluate object grounding on both the Replica and Nr3D datasets. We generated scene graphs using VIZOR for Replica scenes and ScanNet scenes that are included in the test set of Nr3D. Due to the absence of ground truth, we are not able to evaluate the accuracy of the scene graphs directly. Rather, we evaluate them using downstream tasks.

\textbf{Relevant Context Filtering:} For complex scenes with numerous objects, the scene graph can be very large, making direct reasoning inefficient. To address this, we introduce a \textit{scene graph pruning mechanism} that dynamically reduces the graph size based on query relevance, ensuring efficient and focused reasoning. 
This is achieved by embedding both the text query and each subject, predicate, and object triplet using a sentence transformer\footnote{\url{https://huggingface.co/sentence-transformers/all-mpnet-base-v2}}. We then compute the similarity score between the query embedding and each triplet embedding, selecting the top-$K$ most relevant relations. In our experiments, we set $K = 1500$, meaning we retrieve up to 1500 relations per query, however, in some cases, fewer relations may be selected when the number of relations are already low. 
It is to be noted that, we only prune the relationships and pass all the nodes along with their attributes as context to the LLM for querying.
\begin{table}[h!]
\centering
\resizebox{\columnwidth}{!}{%
\begin{tabular}{c|c|cc|cc} \hline
Query Type & \# Queries & CG-CLIP \cite{gu2024conceptgraphs} & CG-LLM \cite{gu2024conceptgraphs} & VIZOR-LLaMA & VIZOR-GPT \\ \hline
Descriptive & 20 & 0.59 & 0.61 & \textbf{0.75} & 0.63 \\
Affordance & 5 & 0.43 & 0.57 &  \textbf{0.8} & \textbf{0.8} \\
Negation & 5 & 0.26 & 0.80 & 0.9 & \textbf{1}  \\
Complex-Spatial & 30 & 0.20 & 0.28 & 0.58 & \textbf{0.67} \\ \hline
Overall & 60 & 0.37 & 0.56 & 0.76 & \textbf{0.78} \\ \hline
\end{tabular}%
}

\caption{\textbf{Comparison of Open Vocabulary Object Grounding from Text Queries on Replica (room0 and office0)}: We measure the top-1 recall. 
CG-CLIP refers to Conceptgraph with CLIP-based retrieval using cosine similarity. CG-LLM
refers to Conceptgraph with an LLM that parses the scene graph and returns the most relevant object. 
VIZOR always uses LLM-based retrieval techniques, and here we assess different LLMs. All values are evaluated by human evaluators.}
\label{tab:obj_grounding_replica}

\end{table}

\textbf{Experimental Result on Replica:} Following \cite{gu2024conceptgraphs}, we evaluate grounding on 2 scenes (room0, office0) from Replica. Since no ground truth is available, we employ 3 human evaluators to assess the retrieved objects and compare the results. \cite{gu2024conceptgraphs} created questions on room0 and office0 scenes and categorized them in 3 types: Descriptive (e.g.,``A brown chair"), Affordance (e.g., ``Something to sit down") and Negation (e.g.,``place to sit but not lie down". While \cite{gu2024conceptgraphs} uses both CLIP and LLM(GPT4o)-based retrieval (CLIP selects the object with the highest similarity to the query’s embedding, LLM goes through the scene graph nodes to identify the object with the most relevant caption), we use only LLM-based retrieval where we provide the pruned scene graph as context to LLM. Table~\ref{tab:obj_grounding_replica} shows that both VIZOR-LLaMA and VIZOR-GPT outperform \cite{gu2024conceptgraphs}, with VIZOR-GPT achieving a \textbf{22\%} improvement in overall grounding accuracy.
Complex-spatial queries are discussed in Section \ref{sec:complx_query}. 
 
\textbf{Experimental Result on Nr3D:} To show the generalization, we perform object grounding experiments on Nr3D dataset and compare our method against fully supervised \cite{achlioptas2020referit3d, chen2020scanrefer, yuan2021instancerefer, zhao20213dvg, luo20223d, chen2022language}, and state-of-the-art zero-shot methods ZS-3DVG \cite{yuan2024visual}, SeeGround \cite{Li_2025_CVPR}, VLM-Grounder \cite{xuvlm}. Following ZS-3DVG \cite{yuan2024visual}, we use ground-truth object proposals (i.e., we directly use instance segments as nodes and build scene graph by generating node attributes and relationships). 
We compare VIZOR with all state-of-the-art methods on the Nr3D dataset in Table \ref{tab:sota_scannet_compare}. 
Our training-free model performs better or have comparable performance to supervised methods. Also, VIZOR outperforms the state-of-the-art zero-shot methods with a \textbf{4.81}\% overall improvement. Exciting results on Nr3D provide evidence that the performance of our model is not only restricted to one dataset but spans across different benchmark datasets.

\begin{table}[h!]
\centering
\resizebox{0.95\columnwidth}{!}{%
\begin{tabular}{c|c|ccccc} 
\hline
\multirow{2}{*}{\textbf{Method}} & \multirow{2}{*}{\textbf{Type}} & \multicolumn{5}{c}{\textbf{Nr3D}} \\ 
\cline{3-7}
 &  & \textbf{Overall} & \textbf{Easy} & \textbf{Hard} & \begin{tabular}[c]{@{}c@{}}\textbf{View}\\\textbf{ Dep}\end{tabular} & \begin{tabular}[c]{@{}c@{}}\textbf{View}\\\textbf{ Indep}\end{tabular} \\ 
\hline
\multicolumn{1}{c|}{ReferIt3D \cite{achlioptas2020referit3d}} & \multirow{10}{*}{Supervised} & 35.6 & 43.6 & 27.9 & 32.5 & 37.1 \\
{ScanRefer \cite{chen2020scanrefer}} &  & 34.2 & 41.0 & 23.5 & 29.9 & 35.4 \\
{TGNN \cite{huang2021text}} &  & 37.3 & 44.2 & 30.6 & 35.8 & 38.0 \\
{InstanceRefer \cite{yuan2021instancerefer}} &  & 38.8 & 46.0 & 31.8 & 43.5 & 41.9 \\
{3DVG-Trans \cite{zhao20213dvg}} &  & 40.8 & 48.5 & 34.8 & 44.3 & 43.7 \\
{TransRefer3D \cite{he2021transrefer3d}} &  & 42.1 & 48.5 & 36.0 & 45.4 & 44.9 \\
{LanguageRefer \cite{roh2022languagerefer}} &  & 43.9 & 51.0 & 36.6 & 47.7 & 45.0 \\
{SAT \cite{yang2021sat}} &  & 49.2 & 56.3 & 42.4 & 54.0 & 50.4 \\
{3D-SPS \cite{luo20223d}} &  & 51.5 & 58.1 & 45.1 & 55.4 & 53.2 \\
{ViL3DRel \cite{chen2022language}} &  & 64.4 & 70.2 & 57.4 & 62.0 & 64.5 \\ \hline
{ZS-3DVG \cite{yuan2024visual}} & {\multirow{4}{*}{Zero-shot}} & 39.0 & 46.5 & 31.7 & 36.8 & 40.0 \\
{SeeGround \cite{Li_2025_CVPR}} & \multicolumn{1}{c|}{} & 46.1 & 54.8 & 38.3 & 42.3 & 48.2 \\
{VLM-Grounder \cite{xuvlm}} & \multicolumn{1}{c|}{} & \underline{48.0} & \underline{55.2} & \underline{39.5} & \textbf{45.8} & \underline{49.4} \\
{VIZOR-GPT (Ours)} & \multicolumn{1}{c|}{} & \textbf{52.81} & \textbf{62.52 }& \textbf{43.48} & \underline{43.02} &\textbf{ 57.66} \\ \hline
\end{tabular}%
}
\caption{\textbf{Comparison of SoTA method on Grounding Accuracy (\%) on Nr3D dataset} with ground-truth object proposals.}
\label{tab:sota_scannet_compare}

\end{table}
 A detailed comparative discussion of different zero-shot methods including VIZOR is available in supplementary.

\subsection{Complex Visual-Language Queries} \label{sec:complx_query}
While the previous methods are evaluated on object grounding based on text queries, we extend this evaluation further. For room0 and office0 scenes from Replica, we carefully construct complex visual-language queries that require spatial understanding of the scene and additional reasoning to retrieve the object from the scene. We create 30 questions per scene. A list of all the questions can be found in the supplementary. Example includes: ``Where can I sit so that I can read a book at night and I want to face the cabinet while sitting?", ``What is to the 45 degree left side of the screens?". 
Table~\ref{tab:obj_grounding_replica} (4th row) compares our method with CG-CLIP and CG-LLM. Despite CG-LLM also using GPT-4o, VIZOR-GPT achieves a \textbf{39}\% improvement, demonstrating superior scene understanding and reasoning via structured graphs and attribute-rich nodes.

\subsection{Qualitative Analysis}
\begin{figure*}[ht!]
    \centering 
    \includegraphics[width=0.95\textwidth, trim= 0cm 15.3cm 0cm 0.5cm]{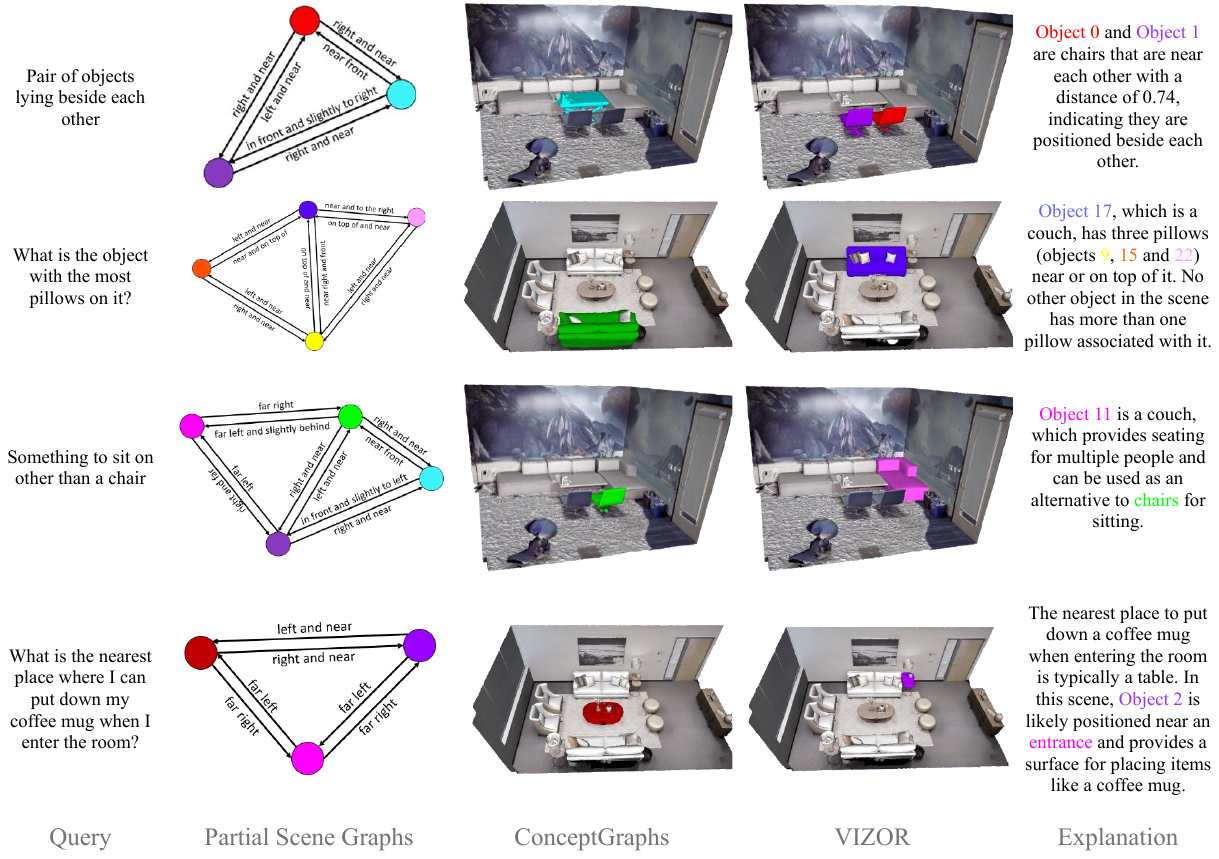}%
    \caption{\textbf{Qualitative Results:} (a) Input query, (b) Part of scene graph generated by VIZOR, (c) ConceptGraphs \cite{gu2024conceptgraphs},  (d) VIZOR output, (e) Explanation of object grounding using VIZOR.}
    \label{fig:qualitative}
    \vspace{-0.5cm}
\end{figure*}

\textbf{Analysis on Replica:} We qualitatively compare VIZOR with ConceptGraphs~\cite{gu2024conceptgraphs} on the task of open vocabulary object grounding using complex natural language queries. Figure~\ref{fig:qualitative} shows results for 4 queries selected from Sec~\ref{sec:complx_query}. For analysis purposes, we show partial 3D scene graph generated by VIZOR in 2nd column. This graph contains the color-coded nodes required to answer the given query, along with relations between them. 
We observe that \cite{gu2024conceptgraphs} fails in answering complex queries (3rd column). Since ConceptGraphs uses CLIP \cite{radford2021learning} similarity score to answer the query, it directly responds with the object mentioned in the query instead of understanding the semantics (Row 3). In contrast, VIZOR correctly recognizes objects even with complex queries, which is visible in the 4th column. An explanation for object grounding by VIZOR is displayed in the last column. It is also noted that, based on the query, VIZOR can refer to multiple objects as per requirement (Row 1). These results indicate that VIZOR is capable of grounding objects by understanding the semantics of the input query and input scene. More examples, along with scene graphs on ScanNet \cite{dai2017scannet}, are shown in supplementary. 

\begin{figure}[h!]
    \centering
    \includegraphics[width=0.94\columnwidth, trim= 0cm 15.5cm 0.5cm 1.2cm]{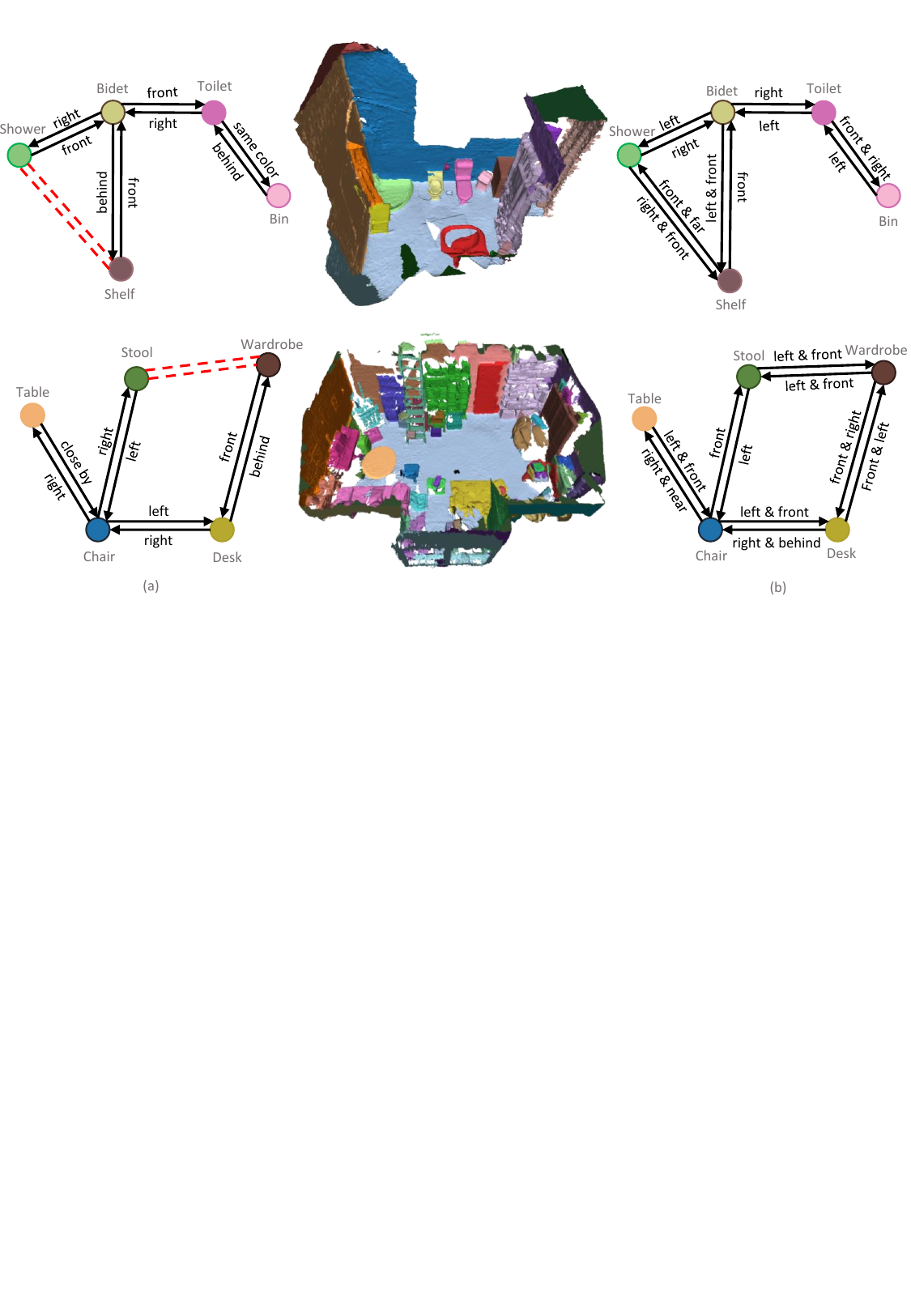}
    \caption{\textbf{Comparative Scene Graph Results:} Scene graph comparison between (a) 3DSSG ground truth scene graphs and (b) VIZOR generated scene graphs on 3DSSG scenes}
    \label{fig:qualitative_3dssg}
    \vspace{-0.2cm}
\end{figure}

\textbf{Analysis on 3DSSG:} We evaluate quality of scene graphs generated by VIZOR on 3DSSG dataset \cite{3DSSG2020}, comparing them against ground truth annotations (Fig.~\ref{fig:qualitative_3dssg}). We present two different scenes from 3DSSG, along with (a) ground truth 3DSSG scene graph and (b) scene graph by VIZOR. We show partial scene graphs for ease of understanding. The nodes are color-coded with same color as the objects, as in the input scenes. In both examples, we see more spatially accurate relationships provided by VIZOR instead of scene graphs that were annotated from a specific viewpoint in the ground truth. We also see missing relationships in ground truth (marked with red dotted line). However, VIZOR-generated graphs provide complete and spatially consistent set of relationships. We show more scene graphs on 3DSSG using VIZOR in supplementary.

\subsection{Design Choice Analysis} \label{sec:ablation}

We further analyze the impact of key architectural choices in VIZOR. 

\textbf{Front Direction Estimation:} To evaluate front-direction prediction, we compare several strategies: ViT-based cosine similarity, Direct query with MLLMs without reference front-view images, and LLMs with diffusion-generated views as reference images. As shown in Table~\ref{tab:front_direction_comparison}, our method achieves the highest front-view precision. \begin{table}[h!]
\centering
\resizebox{\columnwidth}{!}{%
\begin{tabular}{l|c|c|c}
\hline
\textbf{Method} & \textbf{Diffusion} & \textbf{Reference} & \textbf{Front View Precision (\%)} \\
\hline
ViT-based cosine similarity & \xmark & \cmark & 82.35 \\
Direct multimodal-LLM Query & \xmark & \xmark & 80.22 \\
LLM + 2D diffusion reference View & \cmark & \xmark & 58.82 \\
VIZOR & \xmark & \cmark & \textbf{85.71} \\
\hline
\end{tabular}%
}

\caption{Comparison of front-direction inference methods.}
\label{tab:front_direction_comparison}
\end{table}
\vspace{-0.3cm}

\textbf{Number of Views Used:} We also vary number of rendered viewpoints used for front-view prediction and observe trade-off between front view precision, token cost, and runtime (Table~\ref{tab:supp_viewpoints}). We select balanced configuration that ensures both efficiency and performance. We randomly choose 38 objects from room0 and office0 of Replica \cite{replica19arxiv} and perform the experiments on front direction design choice analysis (Table \ref{tab:front_direction_comparison} and \ref{tab:supp_viewpoints}).
\begin{table}[h!]
\centering
\resizebox{0.7\columnwidth}{!}{%
\begin{tabular}{c|c|c|c} 
\hline
\# Views & Front View Precision (\%) & Avg Time (Sec.) & Avg Tokens \\ 
\hline
8 & 82.36 & 4.02 & 1045 \\
\textbf{12} &\textbf{ 85.71} &\textbf{ 6.67} & \textbf{1387} \\
16 & 86.24 & 10.03 & 1726 \\
20 & 86.92 & 14.21 & 2081 \\
\hline
\end{tabular}%
}
\caption{Effect of number of viewpoints on Front-View Detection}
\label{tab:supp_viewpoints}
\end{table}
\vspace{-0.3cm}

\textbf{Attribute Extractor:} We evaluate node attribute quality across four VIZOR variants. We ask three evaluators to rate the attributes on a 1–5 Likert scale (1=``Poor", 5=``Excellent"). The details are provided in supplementary. We test two MLLMs for attribute extraction: Llama 3.2 11B Vision-Instruct \cite{dubey2024llama3} and GPT4-o \cite{achiam2023gpt}. These variants are denoted as VIZOR-LLaMA and VIZOR-GPT. We evaluate both under two settings: attribute extraction using only the front-view (Single-View) or aggregating attributes from multiple viewpoints (Multi-View) as described in Sec \ref{sec:attributes}. Table~\ref{tab:node_attr_rating} shows that VIZOR-GPT with Multi-View aggregation achieves the highest ratings for both color and geometry extraction, and is thus used in all subsequent experiments. 
\begin{table}[h!]
\centering
\resizebox{\columnwidth}{!}{%
\begin{tabular}{c|cc|cc|cc|cc} 
\hline
\multirow{2}{*}{Scene} & \multicolumn{2}{c|}{\begin{tabular}[c]{@{}c@{}}VIZOR-LLaMA\\Single View\end{tabular}} & \multicolumn{2}{c|}{\begin{tabular}[c]{@{}c@{}}VIZOR-LLaMA\\Multi View\end{tabular}} & \multicolumn{2}{c|}{\begin{tabular}[c]{@{}c@{}}VIZOR-GPT \\Single View\end{tabular}} & \multicolumn{2}{c}{\begin{tabular}[c]{@{}c@{}}VIZOR-GPT \\Multi View\end{tabular}} \\ 
\cline{2-9}
 & Color & \multicolumn{1}{l|}{Geometry} & Color & \multicolumn{1}{l|}{Geometry} & Color & Geometry & Color & Geometry \\ 
\hline
room0 & \textcolor{blue}{\textbf{3.68}} & 2.39 & 3.44 & 2.11 & 3.39 & 3.47 & 3.47 & \textbf{4.11} \\
room1 & 3.69 & 2.42 & 3.23 & 2.57 & 4 & 4.42 & \textcolor{blue}{\textbf{4.03}} & \textbf{4.58} \\
room2 & 3.75 & 2.5 & 3.85 & 2.6 & \textcolor{blue}{\textbf{3.9}} & 3.65 & 3.8 & \textbf{3.7} \\
office0 & \textcolor{blue}{\textbf{4.07}} & 2.07 & 3.93 & 2.29 & 3.93 & 4 & 3.93 & \textbf{4.71} \\
office1 & 2.3 & 1.6 & \textcolor{blue}{\textbf{4}} & 1.5 & 3.2 & 3.7 & 3.3 & \textbf{4} \\
office2 & 4.03 & 2.41 & 3.53 & 2.5 & \textcolor{blue}{\textbf{4.03}} & 4.34 & 3.91 & \textbf{4.40} \\
office3 & 3.5 & 2.19 & 3.75 & 2.31 & 3.94 & \textbf{4.69} & \textcolor{blue}{\textbf{4.06}} & 4.31 \\
office4 & 3.95 & 2.59 & 3.69 & 2.38 & 4.18 & 4.10 & \textcolor{blue}{\textbf{4.41}} & \textbf{4.18} \\ 
\hline
Average & 3.62 & 2.27 & 3.68 & 2.28 & 3.82 & 4.05 & \textcolor{blue}{\textbf{3.86}} & \textbf{4.25} \\
\hline
\end{tabular}%
}
\caption{\textbf{Rating of Node Attribute on Replica:} Color refers to the color of the object and Geometry refers to the geometric appearance of the object. \textcolor{blue}{\textbf{Blue}} refers to best performance in color and \textbf{Black} refers to best performance in geometry.}
\label{tab:node_attr_rating}
\end{table}
\section{Failure Analysis}
\begin{figure}[t]

    \centering
    \includegraphics[width=0.67\columnwidth, trim= 0cm 2cm 0cm 1.5cm]{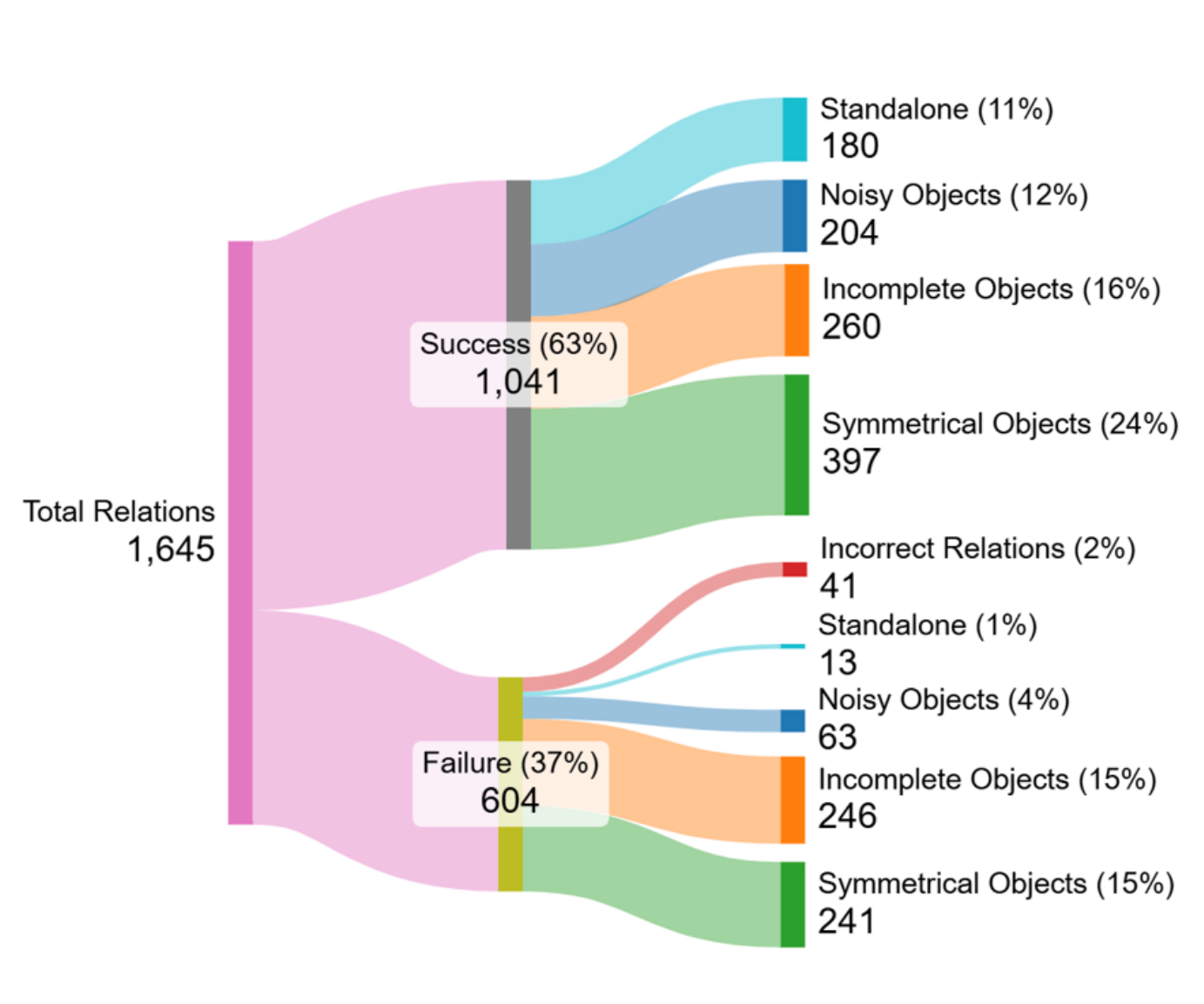}%

    \caption{\textbf{Failure cases: } A graphical representation of failure cases in a hierarchical structure.
    }
    \vspace{-0.5cm}
    \label{fig:failure}
\end{figure}

After thorough experimentation, it is evident that VIZOR can outperform the state-of-the-art methods. Still, there is scope for improvement in the accuracy of our model. We performed extensive analysis on failure cases to understand the bottlenecks of our method in predicting the relations. Our findings are represented in Figure \ref{fig:failure}.

For this failure analysis, we considered four scenes from Replica and four scenes from the Nr3D datasets, comprising a total of 1,645 relations. We conducted a human evaluation to assess the accuracy of these relations. The method can predict the correct relations in case of many symmetrical, noisy, and incomplete objects, but still fails to predict accurate relations for some symmetrical and incomplete objects. The failure of symmetrical objects is due to multiple possible front views that are at equal distance from the scene center, and our model chooses one of them, which may not be the best front view of that object. We observe a failure rate of 37\% ($241$ out of $638$ cases consisting of symmetrical objects) when employing our method (Sec \ref{sec:frontdir}). Similarly, the multi-modal LLM fails to recognize the front view of incomplete objects as it is unable to predict the geometry of that object. The failure rate is very less for standalone relations and noisy objects. Detailed failure analysis and limitations are explained in the supplementary.
\vspace{-0.1in}
\section{Conclusion}
We introduce VIZOR, a training-free method for generating view-invariant scene graphs for 3D scene reasoning. Given a 3D input scene, VIZOR can generate a scene graph containing detailed relations between objects. To the best of our knowledge, we are the first ones to propose a zero-shot approach for scene graph generation based on objects' perspective. We compare our proposed method with existing approaches using qualitative, and quantitative metrics and showcase the proficiency of our method over others.

\noindent\textbf{Acknowledgments:} This work is supported by Fujitsu Research India Private Limited. We thank the evaluators for participating in user study evaluation.

{\small
\bibliographystyle{ieee_fullname}
\bibliography{egbib}

@INPROCEEDINGS{Schult23ICRA,
  author={Schult, Jonas and Engelmann, Francis and Hermans, Alexander and Litany, Or and Tang, Siyu and Leibe, Bastian},
  booktitle={2023 IEEE International Conference on Robotics and Automation (ICRA)}, 
  title={Mask3D: Mask Transformer for 3D Semantic Instance Segmentation}, 
  year={2023},
  volume={},
  number={},
  pages={8216-8223},
  doi={10.1109/ICRA48891.2023.10160590}}

@article{jun2023shap,
  title={Shap-e: Generating conditional 3d implicit functions},
  author={Jun, Heewoo and Nichol, Alex},
  journal={arXiv preprint arXiv:2305.02463},
  year={2023}
}

@article{dubey2024llama3,
  title={The llama 3 herd of models},
  author={Dubey, Abhimanyu and Jauhri, Abhinav and Pandey, Abhinav and Kadian, Abhishek and Al-Dahle, Ahmad and Letman, Aiesha and Mathur, Akhil and Schelten, Alan and Yang, Amy and Fan, Angela and others},
  journal={arXiv preprint arXiv:2407.21783},
  year={2024}
}

@article{achiam2023gpt,
  title={Gpt-4 technical report},
  author={Achiam, Josh and Adler, Steven and Agarwal, Sandhini and Ahmad, Lama and Akkaya, Ilge and Aleman, Florencia Leoni and Almeida, Diogo and Altenschmidt, Janko and Altman, Sam and Anadkat, Shyamal and others},
  journal={arXiv preprint arXiv:2303.08774},
  year={2023}
}

@article{replica19arxiv,
  title =   {The {R}eplica Dataset: A Digital Replica of Indoor Spaces},
  author =  {Julian Straub and Thomas Whelan and Lingni Ma and Yufan Chen and Erik Wijmans and Simon Green and Jakob J. Engel and Raul Mur-Artal and Carl Ren and Shobhit Verma and Anton Clarkson and Mingfei Yan and Brian Budge and Yajie Yan and Xiaqing Pan and June Yon and Yuyang Zou and Kimberly Leon and Nigel Carter and Jesus Briales and  Tyler Gillingham and  Elias Mueggler and Luis Pesqueira and Manolis Savva and Dhruv Batra and Hauke M. Strasdat and Renzo De Nardi and Michael Goesele and Steven Lovegrove and Richard Newcombe },
  journal = {arXiv preprint arXiv:1906.05797},
  year =    {2019}
}

@inproceedings{gu2024conceptgraphs,
  title={Conceptgraphs: Open-vocabulary 3d scene graphs for perception and planning},
  author={Gu, Qiao and Kuwajerwala, Ali and Morin, Sacha and Jatavallabhula, Krishna Murthy and Sen, Bipasha and Agarwal, Aditya and Rivera, Corban and Paul, William and Ellis, Kirsty and Chellappa, Rama and others},
  booktitle={2024 IEEE International Conference on Robotics and Automation (ICRA)},
  pages={5021--5028},
  year={2024},
  organization={IEEE}
}

@inproceedings{achlioptas2020referit3d,
  title={Referit3d: Neural listeners for fine-grained 3d object identification in real-world scenes},
  author={Achlioptas, Panos and Abdelreheem, Ahmed and Xia, Fei and Elhoseiny, Mohamed and Guibas, Leonidas},
  booktitle={Computer Vision--ECCV 2020: 16th European Conference, Glasgow, UK, August 23--28, 2020, Proceedings, Part I 16},
  pages={422--440},
  year={2020},
  organization={Springer}
}

@inproceedings{dai2017scannet,
  title={Scannet: Richly-annotated 3d reconstructions of indoor scenes},
  author={Dai, Angela and Chang, Angel X and Savva, Manolis and Halber, Maciej and Funkhouser, Thomas and Nie{\ss}ner, Matthias},
  booktitle={Proceedings of the IEEE conference on computer vision and pattern recognition},
  pages={5828--5839},
  year={2017}
}

@inproceedings{chen2020scanrefer,
  title={Scanrefer: 3d object localization in rgb-d scans using natural language},
  author={Chen, Dave Zhenyu and Chang, Angel X and Nie{\ss}ner, Matthias},
  booktitle={European conference on computer vision},
  pages={202--221},
  year={2020},
  organization={Springer}
}

@inproceedings{huang2021text,
  title={Text-guided graph neural networks for referring 3d instance segmentation},
  author={Huang, Pin-Hao and Lee, Han-Hung and Chen, Hwann-Tzong and Liu, Tyng-Luh},
  booktitle={Proceedings of the AAAI Conference on Artificial Intelligence},
  volume={35},
  number={2},
  pages={1610--1618},
  year={2021}
}

@inproceedings{yuan2021instancerefer,
  title={Instancerefer: Cooperative holistic understanding for visual grounding on point clouds through instance multi-level contextual referring},
  author={Yuan, Zhihao and Yan, Xu and Liao, Yinghong and Zhang, Ruimao and Wang, Sheng and Li, Zhen and Cui, Shuguang},
  booktitle={Proceedings of the IEEE/CVF International Conference on Computer Vision},
  pages={1791--1800},
  year={2021}
}

@inproceedings{zhao20213dvg,
  title={3DVG-Transformer: Relation modeling for visual grounding on point clouds},
  author={Zhao, Lichen and Cai, Daigang and Sheng, Lu and Xu, Dong},
  booktitle={Proceedings of the IEEE/CVF International Conference on Computer Vision},
  pages={2928--2937},
  year={2021}
}

@inproceedings{he2021transrefer3d,
  title={Transrefer3d: Entity-and-relation aware transformer for fine-grained 3d visual grounding},
  author={He, Dailan and Zhao, Yusheng and Luo, Junyu and Hui, Tianrui and Huang, Shaofei and Zhang, Aixi and Liu, Si},
  booktitle={Proceedings of the 29th ACM International Conference on Multimedia},
  pages={2344--2352},
  year={2021}
}

@inproceedings{yang2021sat,
  title={Sat: 2d semantics assisted training for 3d visual grounding},
  author={Yang, Zhengyuan and Zhang, Songyang and Wang, Liwei and Luo, Jiebo},
  booktitle={Proceedings of the IEEE/CVF International Conference on Computer Vision},
  pages={1856--1866},
  year={2021}
}

@inproceedings{luo20223d,
  title={3d-sps: Single-stage 3d visual grounding via referred point progressive selection},
  author={Luo, Junyu and Fu, Jiahui and Kong, Xianghao and Gao, Chen and Ren, Haibing and Shen, Hao and Xia, Huaxia and Liu, Si},
  booktitle={Proceedings of the IEEE/CVF Conference on Computer Vision and Pattern Recognition},
  pages={16454--16463},
  year={2022}
}

@inproceedings{huang2022multi,
  title={Multi-view transformer for 3d visual grounding},
  author={Huang, Shijia and Chen, Yilun and Jia, Jiaya and Wang, Liwei},
  booktitle={Proceedings of the IEEE/CVF Conference on Computer Vision and Pattern Recognition},
  pages={15524--15533},
  year={2022}
}

@article{chen2022language,
  title={Language conditioned spatial relation reasoning for 3d object grounding},
  author={Chen, Shizhe and Guhur, Pierre-Louis and Tapaswi, Makarand and Schmid, Cordelia and Laptev, Ivan},
  journal={Advances in neural information processing systems},
  volume={35},
  pages={20522--20535},
  year={2022}
}

@inproceedings{3DSSG2020,
    title={Learning 3D Semantic Scene Graphs from 3D Indoor Reconstructions},
    author={Wald, Johanna and Dhamo, Helisa and Navab, Nassir and Tombari, Federico},
    booktitle={Conference on Computer Vision and Pattern Recognition (CVPR)}, 
    year={2020}
  }

@ARTICLE{sc1,
  author={Phueaksri, Itthisak and Kastner, Marc A. and Kawanishi, Yasutomo and Komamizu, Takahiro and Ide, Ichiro},
  journal={IEEE Access}, 
  title={An Approach to Generate a Caption for an Image Collection Using Scene Graph Generation}, 
  year={2023},
  volume={11},
  number={},
  pages={128245-128260},
  doi={10.1109/ACCESS.2023.3332098}}

@InProceedings{sc2,
author = {Yang, Xu and Tang, Kaihua and Zhang, Hanwang and Cai, Jianfei},
title = {Auto-Encoding Scene Graphs for Image Captioning},
booktitle = {Proceedings of the IEEE/CVF Conference on Computer Vision and Pattern Recognition (CVPR)},
month = {June},
year = {2019}
}

@InProceedings{sc3,
author="Zhong, Yiwu
and Wang, Liwei
and Chen, Jianshu
and Yu, Dong
and Li, Yin",
editor="Vedaldi, Andrea
and Bischof, Horst
and Brox, Thomas
and Frahm, Jan-Michael",
title="Comprehensive Image Captioning via Scene Graph Decomposition",
booktitle="Computer Vision -- ECCV 2020",
year="2020",
publisher="Springer International Publishing",
address="Cham",
pages="211--229",
isbn="978-3-030-58568-6"
}

@InProceedings{sc4,
author = {Gu, Jiuxiang and Joty, Shafiq and Cai, Jianfei and Zhao, Handong and Yang, Xu and Wang, Gang},
title = {Unpaired Image Captioning via Scene Graph Alignments},
booktitle = {Proceedings of the IEEE/CVF International Conference on Computer Vision (ICCV)},
month = {October},
year = {2019}
}

@ARTICLE{sc5,
  author={Li, Xiangyang and Jiang, Shuqiang},
  journal={IEEE Transactions on Multimedia}, 
  title={Know More Say Less: Image Captioning Based on Scene Graphs}, 
  year={2019},
  volume={21},
  number={8},
  pages={2117-2130},
  doi={10.1109/TMM.2019.2896516}}

@ARTICLE{rn1,
  author={Bavle, Hriday and Sanchez-Lopez, Jose Luis and Shaheer, Muhammad and Civera, Javier and Voos, Holger},
  journal={IEEE Robotics and Automation Letters}, 
  title={Situational Graphs for Robot Navigation in Structured Indoor Environments}, 
  year={2022},
  volume={7},
  number={4},
  pages={9107-9114},
  doi={10.1109/LRA.2022.3189785}}

@INPROCEEDINGS{rn2,
  author={Blumenthal, Sebastian and Bruyninckx, Herman and Nowak, Walter and Prassler, Erwin},
  booktitle={2013 IEEE International Conference on Robotics and Automation}, 
  title={A scene graph based shared 3D world model for robotic applications}, 
  year={2013},
  volume={},
  number={},
  pages={453-460},
  doi={10.1109/ICRA.2013.6630614}}

@inproceedings{rn3,
  title={Hierarchical open-vocabulary 3d scene graphs for language-grounded robot navigation},
  author={Werby, Abdelrhman and Huang, Chenguang and B{\"u}chner, Martin and Valada, Abhinav and Burgard, Wolfram},
  booktitle={First Workshop on Vision-Language Models for Navigation and Manipulation at ICRA 2024},
  year={2024}
}

@inproceedings{rn4,
  title={Natural language-guided semantic navigation using scene graph},
  author={Kim, Dohyun and Kim, Jinwoo and Cho, Minwoo and Park, Daehyung},
  booktitle={International Conference on Robot Intelligence Technology and Applications},
  pages={148--156},
  year={2022},
  organization={Springer}
}

@INPROCEEDINGS{rn5,
  author={Ni, Zhe and Deng, Xiaoxin and Tai, Cong and Zhu, Xinyue and Xie, Qinghongbing and Huang, Weihang and Wu, Xiang and Zeng, Long},
  booktitle={2024 IEEE/RSJ International Conference on Intelligent Robots and Systems (IROS)}, 
  title={GRID: Scene-Graph-based Instruction-driven Robotic Task Planning}, 
  year={2024},
  volume={},
  number={},
  pages={13765-13772},
  doi={10.1109/IROS58592.2024.10801291}}

@inproceedings{rn6,
  title={Taskography: Evaluating robot task planning over large 3d scene graphs},
  author={Agia, Christopher and Jatavallabhula, Krishna Murthy and Khodeir, Mohamed and Miksik, Ondrej and Vineet, Vibhav and Mukadam, Mustafa and Paull, Liam and Shkurti, Florian},
  booktitle={Conference on Robot Learning},
  pages={46--58},
  year={2022},
  organization={PMLR}
}

@InProceedings{sm1,
    author    = {Dhamo, Helisa and Manhardt, Fabian and Navab, Nassir and Tombari, Federico},
    title     = {Graph-to-3D: End-to-End Generation and Manipulation of 3D Scenes Using Scene Graphs},
    booktitle = {Proceedings of the IEEE/CVF International Conference on Computer Vision (ICCV)},
    month     = {October},
    year      = {2021},
    pages     = {16352-16361}
}

@INPROCEEDINGS{sm2,
  author={Jiao, Ziyuan and Niu, Yida and Zhang, Zeyu and Zhu, Song-Chun and Zhu, Yixin and Liu, Hangxin},
  booktitle={2022 IEEE/RSJ International Conference on Intelligent Robots and Systems (IROS)}, 
  title={Sequential Manipulation Planning on Scene Graph}, 
  year={2022},
  volume={},
  number={},
  pages={8203-8210},
  doi={10.1109/IROS47612.2022.9981735}}

@inproceedings{sm3,
author = {Su, Sitong and Gao, Lianli and Zhu, Junchen and Shao, Jie and Song, Jingkuan},
title = {Fully Functional Image Manipulation Using Scene Graphs in A Bounding-Box Free Way},
year = {2021},
isbn = {9781450386517},
publisher = {Association for Computing Machinery},
address = {New York, NY, USA},
url = {https://doi.org/10.1145/3474085.3475326},
doi = {10.1145/3474085.3475326},
booktitle = {Proceedings of the 29th ACM International Conference on Multimedia},
pages = {1784–1792},
numpages = {9},
location = {Virtual Event, China},
series = {MM '21}
}

@InProceedings{sm4,
author = {Dhamo, Helisa and Farshad, Azade and Laina, Iro and Navab, Nassir and Hager, Gregory D. and Tombari, Federico and Rupprecht, Christian},
title = {Semantic Image Manipulation Using Scene Graphs},
booktitle = {Proceedings of the IEEE/CVF Conference on Computer Vision and Pattern Recognition (CVPR)},
month = {June},
year = {2020}
}

@article{vqa1,
  title={Scene graph refinement network for visual question answering},
  author={Qian, Tianwen and Chen, Jingjing and Chen, Shaoxiang and Wu, Bo and Jiang, Yu-Gang},
  journal={IEEE Transactions on Multimedia},
  volume={25},
  pages={3950--3961},
  year={2022},
  publisher={IEEE}
}

@inproceedings{vqa2,
  title={Visual question answering over scene graph},
  author={Lee, Soohyeong and Kim, Ju-Whan and Oh, Youngmin and Jeon, Joo Hyuk},
  booktitle={2019 First International Conference on Graph Computing (GC)},
  pages={45--50},
  year={2019},
  organization={IEEE}
}

@inproceedings{vqa3,
  title={Lightweight visual question answering using scene graphs},
  author={Nuthalapati, Sai Vidyaranya and Chandradevan, Ramraj and Giunchiglia, Eleonora and Li, Bowen and Kayser, Maxime and Lukasiewicz, Thomas and Yang, Carl},
  booktitle={Proceedings of the 30th ACM International Conference on Information \& Knowledge Management},
  pages={3353--3357},
  year={2021}
}

@inproceedings{vqa4,
  title={Robotvqa—a scene-graph-and deep-learning-based visual question answering system for robot manipulation},
  author={Kenfack, Franklin Kenghagho and Siddiky, Feroz Ahmed and Balint-Benczedi, Ferenc and Beetz, Michael},
  booktitle={2020 IEEE/RSJ International Conference on Intelligent Robots and Systems (IROS)},
  pages={9667--9674},
  year={2020},
  organization={IEEE}
}

@inproceedings{lv2024sgformer,
  title={Sgformer: Semantic graph transformer for point cloud-based 3d scene graph generation},
  author={Lv, Changsheng and Qi, Mengshi and Li, Xia and Yang, Zhengyuan and Ma, Huadong},
  booktitle={Proceedings of the AAAI Conference on Artificial Intelligence},
  volume={38},
  number={5},
  pages={4035--4043},
  year={2024}
}

@inproceedings{koch2024open3dsg,
  title={Open3dsg: Open-vocabulary 3d scene graphs from point clouds with queryable objects and open-set relationships},
  author={Koch, Sebastian and Vaskevicius, Narunas and Colosi, Mirco and Hermosilla, Pedro and Ropinski, Timo},
  booktitle={Proceedings of the IEEE/CVF Conference on Computer Vision and Pattern Recognition},
  pages={14183--14193},
  year={2024}
}

@inproceedings{looper20233d,
  title={3d vsg: Long-term semantic scene change prediction through 3d variable scene graphs},
  author={Looper, Samuel and Rodriguez-Puigvert, Javier and Siegwart, Roland and Cadena, Cesar and Schmid, Lukas},
  booktitle={2023 IEEE International Conference on Robotics and Automation (ICRA)},
  pages={8179--8186},
  year={2023},
  organization={IEEE}
}

@article{rana2023sayplan,
  title={Sayplan: Grounding large language models using 3d scene graphs for scalable robot task planning},
  author={Rana, Krishan and Haviland, Jesse and Garg, Sourav and Abou-Chakra, Jad and Reid, Ian and Suenderhauf, Niko},
  journal={arXiv preprint arXiv:2307.06135},
  year={2023}
}

@inproceedings{zhong2021learning,
  title={Learning to generate scene graph from natural language supervision},
  author={Zhong, Yiwu and Shi, Jing and Yang, Jianwei and Xu, Chenliang and Li, Yin},
  booktitle={Proceedings of the IEEE/CVF International Conference on Computer Vision},
  pages={1823--1834},
  year={2021}
}

@article{liu2022explore,
  title={Explore contextual information for 3d scene graph generation},
  author={Liu, Yuanyuan and Long, Chengjiang and Zhang, Zhaoxuan and Liu, Bokai and Zhang, Qiang and Yin, Baocai and Yang, Xin},
  journal={IEEE Transactions on Visualization and Computer Graphics},
  volume={29},
  number={12},
  pages={5556--5568},
  year={2022},
  publisher={IEEE}
}

@inproceedings{roh2022languagerefer,
  title={Languagerefer: Spatial-language model for 3d visual grounding},
  author={Roh, Junha and Desingh, Karthik and Farhadi, Ali and Fox, Dieter},
  booktitle={Conference on Robot Learning},
  pages={1046--1056},
  year={2022},
  organization={PMLR}
}

@inproceedings{kamath2021mdetr,
  title={Mdetr-modulated detection for end-to-end multi-modal understanding},
  author={Kamath, Aishwarya and Singh, Mannat and LeCun, Yann and Synnaeve, Gabriel and Misra, Ishan and Carion, Nicolas},
  booktitle={Proceedings of the IEEE/CVF international conference on computer vision},
  pages={1780--1790},
  year={2021}
}

@inproceedings{yu2016modeling,
  title={Modeling context in referring expressions},
  author={Yu, Licheng and Poirson, Patrick and Yang, Shan and Berg, Alexander C and Berg, Tamara L},
  booktitle={Computer Vision--ECCV 2016: 14th European Conference, Amsterdam, The Netherlands, October 11-14, 2016, Proceedings, Part II 14},
  pages={69--85},
  year={2016},
  organization={Springer}
}

@inproceedings{yu2018mattnet,
  title={Mattnet: Modular attention network for referring expression comprehension},
  author={Yu, Licheng and Lin, Zhe and Shen, Xiaohui and Yang, Jimei and Lu, Xin and Bansal, Mohit and Berg, Tamara L},
  booktitle={Proceedings of the IEEE conference on computer vision and pattern recognition},
  pages={1307--1315},
  year={2018}
}

@inproceedings{velivckovic2018graph,
  title={Graph Attention Networks},
  author={Veli{\v{c}}kovi{\'c}, Petar and Cucurull, Guillem and Casanova, Arantxa and Romero, Adriana and Li{\`o}, Pietro and Bengio, Yoshua},
  booktitle={International Conference on Learning Representations},
  year={2018}
}

@article{wang2019dynamic,
  title={Dynamic graph cnn for learning on point clouds},
  author={Wang, Yue and Sun, Yongbin and Liu, Ziwei and Sarma, Sanjay E and Bronstein, Michael M and Solomon, Justin M},
  journal={ACM Transactions on Graphics (tog)},
  volume={38},
  number={5},
  pages={1--12},
  year={2019},
  publisher={Acm New York, NY, USA}
}

@inproceedings{armeni20193d,
  title={3d scene graph: A structure for unified semantics, 3d space, and camera},
  author={Armeni, Iro and He, Zhi-Yang and Gwak, JunYoung and Zamir, Amir R and Fischer, Martin and Malik, Jitendra and Savarese, Silvio},
  booktitle={Proceedings of the IEEE/CVF international conference on computer vision},
  pages={5664--5673},
  year={2019}
}

@article{hughes2022hydra,
  title={Hydra: A Real-time Spatial Perception Engine for 3D Scene Graph Construction and Optimization},
  author={Hughes, Nathan and Chang, Yun and Carlone, Luca},
  journal={CoRR},
  year={2022}
}

@inproceedings{rosinol20203d,
  title={3D Dynamic Scene Graphs: Actionable Spatial Perception with Places, Objects, and Humans},
  author={Rosinol, Antoni and Gupta, Arjun and Abate, Marcus and Shi, Jingnan and Carlone, Luca},
  booktitle={Robotics: Science and Systems},
  year={2020}
}

@article{rosinol2021kimera,
  title={Kimera: From SLAM to spatial perception with 3D dynamic scene graphs},
  author={Rosinol, Antoni and Violette, Andrew and Abate, Marcus and Hughes, Nathan and Chang, Yun and Shi, Jingnan and Gupta, Arjun and Carlone, Luca},
  journal={The International Journal of Robotics Research},
  volume={40},
  number={12-14},
  pages={1510--1546},
  year={2021},
  publisher={SAGE Publications Sage UK: London, England}
}

@inproceedings{koch2024lang3dsg,
  title={Lang3dsg: Language-based contrastive pre-training for 3d scene graph prediction},
  author={Koch, Sebastian and Hermosilla, Pedro and Vaskevicius, Narunas and Colosi, Mirco and Ropinski, Timo},
  booktitle={2024 International Conference on 3D Vision (3DV)},
  pages={1037--1047},
  year={2024},
  organization={IEEE}
}

@inproceedings{wald2020learning,
  title={Learning 3d semantic scene graphs from 3d indoor reconstructions},
  author={Wald, Johanna and Dhamo, Helisa and Navab, Nassir and Tombari, Federico},
  booktitle={Proceedings of the IEEE/CVF Conference on Computer Vision and Pattern Recognition},
  pages={3961--3970},
  year={2020}
}

@inproceedings{zhang2021exploiting,
  title={Exploiting edge-oriented reasoning for 3d point-based scene graph analysis},
  author={Zhang, Chaoyi and Yu, Jianhui and Song, Yang and Cai, Weidong},
  booktitle={Proceedings of the IEEE/CVF conference on computer vision and pattern recognition},
  pages={9705--9715},
  year={2021}
}

@article{zhang2021knowledge,
  title={Knowledge-inspired 3d scene graph prediction in point cloud},
  author={Zhang, Shoulong and Hao, Aimin and Qin, Hong and others},
  journal={Advances in Neural Information Processing Systems},
  volume={34},
  pages={18620--18632},
  year={2021}
}

@article{liu2023visual,
  title={Visual instruction tuning},
  author={Liu, Haotian and Li, Chunyuan and Wu, Qingyang and Lee, Yong Jae},
  journal={Advances in neural information processing systems},
  volume={36},
  pages={34892--34916},
  year={2023}
}

@inproceedings{lin2023instructscene,
  title={InstructScene: Instruction-Driven 3D Indoor Scene Synthesis with Semantic Graph Prior},
  author={Lin, Chenguo and Yadong, MU},
  booktitle={The Twelfth International Conference on Learning Representations},
  year={2023}
}

@article{feng2023layoutgpt,
  title={Layoutgpt: Compositional visual planning and generation with large language models},
  author={Feng, Weixi and Zhu, Wanrong and Fu, Tsu-jui and Jampani, Varun and Akula, Arjun and He, Xuehai and Basu, Sugato and Wang, Xin Eric and Wang, William Yang},
  journal={Advances in Neural Information Processing Systems},
  volume={36},
  pages={18225--18250},
  year={2023}
}

@inproceedings{yuan2024visual,
  title={Visual programming for zero-shot open-vocabulary 3d visual grounding},
  author={Yuan, Zhihao and Ren, Jinke and Feng, Chun-Mei and Zhao, Hengshuang and Cui, Shuguang and Li, Zhen},
  booktitle={Proceedings of the IEEE/CVF Conference on Computer Vision and Pattern Recognition},
  pages={20623--20633},
  year={2024}
}

@misc{linok2024barequeriesopenvocabularyobject,
      title={Beyond Bare Queries: Open-Vocabulary Object Grounding with 3D Scene Graph}, 
      author={Sergey Linok and Tatiana Zemskova and Svetlana Ladanova and Roman Titkov and Dmitry Yudin and Maxim Monastyrny and Aleksei Valenkov},
      year={2024},
      eprint={2406.07113},
      archivePrefix={arXiv},
      primaryClass={cs.CV},
      url={https://arxiv.org/abs/2406.07113}, 
}

@article{yang2023sam3d,
  title={Sam3d: Segment anything in 3d scenes},
  author={Yang, Yunhan and Wu, Xiaoyang and He, Tong and Zhao, Hengshuang and Liu, Xihui},
  journal={arXiv preprint arXiv:2306.03908},
  year={2023}
}

@inproceedings{
takmaz2023openmaskd,
title={OpenMask3D: Open-Vocabulary 3D Instance Segmentation},
author={Ay{\c{c}}a Takmaz and Elisabetta Fedele and Robert Sumner and Marc Pollefeys and Federico Tombari and Francis Engelmann},
booktitle={Thirty-seventh Conference on Neural Information Processing Systems},
year={2023},
url={https://openreview.net/forum?id=8vuDHCxrmy}
}

@InProceedings{Li_2025_CVPR,
    author    = {Li, Rong and Li, Shijie and Kong, Lingdong and Yang, Xulei and Liang, Junwei},
    title     = {SeeGround: See and Ground for Zero-Shot Open-Vocabulary 3D Visual Grounding},
    booktitle = {Proceedings of the Computer Vision and Pattern Recognition Conference (CVPR)},
    month     = {June},
    year      = {2025},
    pages     = {3707-3717}
}

@inproceedings{xuvlm,
  title={VLM-Grounder: A VLM Agent for Zero-Shot 3D Visual Grounding},
  author={Xu, Runsen and Huang, Zhiwei and Wang, Tai and Chen, Yilun and Pang, Jiangmiao and Lin, Dahua},
  booktitle={8th Annual Conference on Robot Learning},
  year={2024}
}

@inproceedings{Wald_2020_3RScan,
  title     = {3RScan: A Reality-Based 3D Dataset for Scene Understanding},
  author    = {Wald, Johannes and Dahnert, Markus and R{\"u}ckert, Daniel and Savva, Manolis and Rother, Carsten and Tombari, Federico and Navab, Nassir},
  booktitle = {Proceedings of the IEEE/CVF Conference on Computer Vision and Pattern Recognition (CVPR)},
  year      = {2020},
  pages     = {8180--8189}
}

@inproceedings{zhang2024recognize,
  title={Recognize anything: A strong image tagging model},
  author={Zhang, Youcai and Huang, Xinyu and Ma, Jinyu and Li, Zhaoyang and Luo, Zhaochuan and Xie, Yanchun and Qin, Yuzhuo and Luo, Tong and Li, Yaqian and Liu, Shilong and others},
  booktitle={Proceedings of the IEEE/CVF Conference on Computer Vision and Pattern Recognition},
  pages={1724--1732},
  year={2024}
}

@inproceedings{liu2024grounding,
  title={Grounding dino: Marrying dino with grounded pre-training for open-set object detection},
  author={Liu, Shilong and Zeng, Zhaoyang and Ren, Tianhe and Li, Feng and Zhang, Hao and Yang, Jie and Jiang, Qing and Li, Chunyuan and Yang, Jianwei and Su, Hang and others},
  booktitle={European Conference on Computer Vision},
  pages={38--55},
  year={2024},
  organization={Springer}
}

@inproceedings{radford2021learning,
  title={Learning transferable visual models from natural language supervision},
  author={Radford, Alec and Kim, Jong Wook and Hallacy, Chris and Ramesh, Aditya and Goh, Gabriel and Agarwal, Sandhini and Sastry, Girish and Askell, Amanda and Mishkin, Pamela and Clark, Jack and others},
  booktitle={International conference on machine learning},
  pages={8748--8763},
  year={2021},
  organization={PmLR}
}

@INPROCEEDINGS{vppnet,
  author={Shi, Xiangxi and Wu, Zhonghua and Lee, Stefan},
  booktitle={2024 IEEE/CVF Conference on Computer Vision and Pattern Recognition (CVPR)}, 
  title={Viewpoint-Aware Visual Grounding in 3D Scenes}, 
  year={2024},
  volume={},
  number={},
  pages={14056-14065},
  doi={10.1109/CVPR52733.2024.01333}}

@article{shang2022learning,
  title={Learning viewpoint-agnostic visual representations by recovering tokens in 3d space},
  author={Shang, Jinghuan and Das, Srijan and Ryoo, Michael},
  journal={Advances in Neural Information Processing Systems},
  volume={35},
  pages={31031--31044},
  year={2022}
}
}

\end{document}